\documentclass[sigconf,screen,nonacm]{acmart}
\usepackage{xspace}
\usepackage{xcolor}

\newcommand{\sysname}{\textsc{ARBot}\xspace} 
\graphicspath{{./}}
\begin{document}

\hypersetup{
  linkcolor=magenta,   citecolor=magenta,
  urlcolor=magenta
}

\title{\sysname: A High-Fidelity Robotic Manipulator Teleoperation Framework for Human-Centered Augmented Reality Evaluation}

\author{Harsh Chhajed}
\affiliation{\institution{Worcester Polytechnic Institute}
  \city{Worcester}
  \state{MA}
  \country{USA}
}
\email{hchhajed@wpi.edu}

\author{Tian Guo}
\affiliation{\institution{Worcester Polytechnic Institute}
  \city{Worcester}
  \state{MA}
  \country{USA}
}
\email{tian@wpi.edu}

\begin{abstract}
Validating Augmented Reality (AR) tracking and interaction models requires precise, repeatable ground-truth motion. 
However, human users cannot reliably perform consistent motion due to biomechanical variability. Robotic manipulators are promising to act as human motion proxies if they can mimic human movements. In this work, we design and implement \sysname, a real-time teleoperation platform that can effectively capture natural human motion and accurately replay the movements via robotic manipulators. \sysname includes two capture models: stable wrist motion capture via a custom CV and IMU pipeline, and natural 6-DOF control via a mobile application. We design a proactively-safe QP controller to ensure smooth, jitter-free execution of the robotic manipulator, enabling it to function as a high-fidelity record and replay physical proxy. We open-source \sysname and release a benchmark dataset of 132 human and synthetic trajectories  captured using \sysname to support controllable and scalable AR evaluation.
\end{abstract}



\maketitle
\section{Introduction}

\begin{figure}[t]
    \centering
    \includegraphics[width=0.8\linewidth]{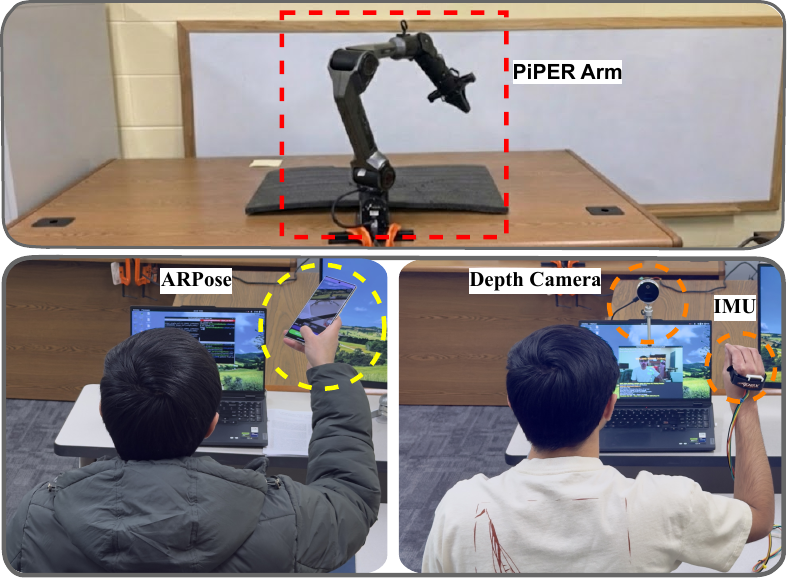}
    \caption{\sysname in Action.
    \textnormal{\sysname serves as a physical proxy for AR evaluation, capable of capturing natural human motion via mobile ARPose (left) or CV+IMU interfaces (right) and replaying it robotic manipulators (top).}}
    \Description{A composite image showing three components of the ARBot system. On the left, a user holds a smartphone running the ARPose application. On the right, a user's hand is shown with a wrist-mounted IMU sensor. At the top, a robotic manipulator arm is depicted executing a trajectory.}
    \label{fig:teaser}
\end{figure}

The rapid advancement of Augmented Reality (AR) has necessitated robust methods for evaluating spatial tracking, rendering latency, and 3D interaction techniques. As AR systems transition from novelties to critical tools for remote inspection and navigation, they rely heavily on the precise correspondence between physical motion and virtual overlays. However, evaluating these systems remains a significant challenge, particularly when trying to isolate the performance of tracking algorithms from the erratic nature of the physical world.

A primary gap in current evaluation methodology is the frequent disconnect between quantitative metrics and the actual user experience. For example, ARCADE~\cite{arcade} demonstrated that computer vision models with identical numerical error scores (e.g., RMSE) can yield vastly different user perceptual qualities in real-world AR tasks~\cite{arcade}. This disconnect necessitates human-centered user studies to capture the nuances of immersion and interaction quality that raw numbers miss.

However, conducting reliable and scalable human-centered evaluation is challenging. Human testers cannot reliably reproduce the same motion across multiple trials due to biomechanical variability, fatigue, and natural tremors. This stochasticity makes it nearly impossible to distinguish whether a tracking failure was caused by the algorithm or an erratic user movement. It is also impractical to have human users repetitively test out different baselines.  
ExpAR envisions addressing this limitation through standardized, programmable mobility platforms~\cite{expar}; however, existing ExpAR prototypes cannot capture and replay AR user hand movements.

In this work, we present \sysname, a proactively-safe robotic manipulator platform designed to serve as the physical proxy for natural human AR interactions. Unlike traditional teleoperation systems that focus solely on task completion, \sysname is designed and engineered to effectively capture natural human motion and accurately replay the motion via a robotic manipulator. 
Figure~\ref{fig:teaser} illustrates the main hardware components of \sysname.

We design \sysname as a three-stage pipeline to ensure both capturing fidelity and replay safety.
First, the \textbf{capture} stage utilizes multimodal interfaces including a touchless Computer Vision and IMU (CV+IMU) tracker and the custom ARPose Android application to record natural human motions. Second, the \textbf{control} stage processes these inputs through a Quadratic Programming (QP) controller that acts as a proactive safety filter, ensuring smooth execution without violating hardware limits. Finally, the \textbf{replay} stage drives the robotic arm to execute the target trajectory, mimicking human dexterity while maintaining the stability required for rigorous testing.

We evaluate \sysname's performance with an IRB-approved user study of 11 participants. Our user study shows that \sysname can accurately mimic user motions in real-time, with about 5.0 mm median absolute trajectory error and 19.5 ms end-to-end latency. The repeatability evaluation also shows that inter-trial variability of human motions can be 10X that of \sysname's robotic manipulator. Moreover, 7 out of 11 participants preferred the ARPose capturing method, while the remaining 4 participants appreciate the hands-free operation of the CV+IMU method. 

\sysname can benefit the research community in several different ways. First, researchers can use \sysname's end-to-end pipeline to conduct reproducible human-centered AR evaluation. Second, the robotic manipulator can engage in autopilot mode to trace standardized geometric trajectories (e.g., circles, squares) with sub-centimeter precision for reproducible benchmarking without human input. Third, in the absence of the robotic manipulator, researchers can use our capture module to accurately collect user motions. 
Fourth, researchers can directly leverage our human hand trajectories dataset that is captured with \sysname. This dataset contains over 132 shape tracing trajectories, which can be used for evaluating various tracking algorithm evaluations such as hand tracking and AR interactions. 

In summary, we make the following contributions:
\begin{itemize}
    \item \textbf{End-to-End Architecture:} We design and implement an end-to-end teleoperation framework \sysname that integrates multimodal motion capture, a proactive safety controller, and a precise replay mechanism. 
    \item \textbf{Human-Centered Evaluation:} We evaluate the usefulness of \sysname as a physical proxy to mimic AR user motions with an IRB-approved user study.
\item \textbf{Open Source Artifacts\footnote{\href{https://github.com/cake-lab/ARBot}{https://github.com/cake-lab/ARBot}}:} We release the full platform codebase and a benchmark dataset containing 132 trajectories of shapes like circle, square, etc captured using both ARPose and CV+IMU methods. \end{itemize}

\section{Related Work}

\textbf{AR Evaluation Frameworks and Metrics.}
As Augmented Reality (AR) systems grow more complex, evaluation has shifted from ad-hoc testing to standardized frameworks. Early efforts often relied on manual logging or isolated component tests, which fail to capture system-wide behavior.

To address this, ILLIXR~\cite{9668280} introduced the first open-source testbed covering the entire XR pipeline from SLAM to rendering, enabling researchers to benchmark system-level metrics like power and motion-to-photon latency. Building on this need for rigorous workflows, ExpAR~\cite{expar} and ARFlow~\cite{10.1145/3638550.3643617} focused on experimentation infrastructure. ExpAR automates the instrumentation of AR apps to log performance data, while ARFlow streamlines the management of high-volume multimodal streams (RGB, Depth, IMU) to ensure reproducible data collection.

However, raw performance metrics often fail to capture the user experience. ARCADE~\cite{arcade} demonstrated that two models with identical error rates can feel vastly different to a user, arguing that evaluation must account for specific failure modes like jitter or drift. PredART~\cite{10.1145/3551349.3561160} took this further by proposing an AI Oracle approach, training neural networks to predict human quality ratings from screenshots, effectively attempting to automate the human judge.

While these works provide robust \textit{software testbeds}~\cite{9668280,expar} and \textit{methodology}~\cite{arcade,10.1145/3551349.3561160}, they all face a common bottleneck: the input source. They rely on either pre-recorded datasets, which lack interactivity, or live human testers, who introduce biomechanical variability. Our ARBot platform bridges this gap by serving as a \textit{deterministic physical input}. It provides the repeatable motion ground truth that frameworks like ILLIXR need to benchmark performance and oracles like PredART need to validate their predictions.

\textbf{Visual Analytics for AR Systems.}
Beyond numerical benchmarks, understanding the why behind an AR failure often requires deep visual inspection.
ARGUS~\cite{argus} introduced a visual analytics framework to debug AI-assisted AR tasks. Instead of just reporting a failure rate, ARGUS temporally aligns sensor streams (gaze, hand tracking) with model predictions to reveal subtle issues like an object detection model working perfectly but the guidance arrow appearing with a 200ms delay.
While ARGUS offers powerful post-hoc analysis, it is limited by the data fed into it. A researcher analyzing a tracking failure during a head turn cannot easily determine if the glitch was caused by the algorithm or the user's specific erratic motion. ARBot complements ARGUS by enabling unit tests for physical motion. Researchers can program the robot to execute precise edge-case trajectories, using ARGUS to visualize exactly how the system state degrades under controlled stress, isolating the algorithm's performance from human inconsistency.

\textbf{Robotic Telepresence as Physical Proxies.}
Robotic systems have been used to bridge the physical-virtual divide, but typically with different goals than evaluation.
VRoxy~\cite{10.1145/3586183.3606743} uses robotic proxies to enable remote collaboration, allowing users to embody a robot in a larger space than their physical room permits. The priority here is \textit{presence} making the interaction feel natural for the operator. Similarly, research in Human-Robot Collaboration (HRC)~\cite{boguslavskii2025} focuses on safety and seamless coordination between humans and machines in shared workspaces.
These systems optimize for \textit{usability} and \textit{embodiment}. In contrast, our work treats the robotic manipulator as a \textit{Scientific Instrument}. Rather than focusing on how well a human can drive the robot, we prioritize \textit{fidelity} and \textit{repeatability}. This shift allows us to generate the high-precision ground truth data required to rigorously evaluate next-generation AR algorithms, complementing the software-focused contributions of ExpAR and ARCADE.

\begin{figure*}[t]
    \centering
    \includegraphics[width=0.8\textwidth]{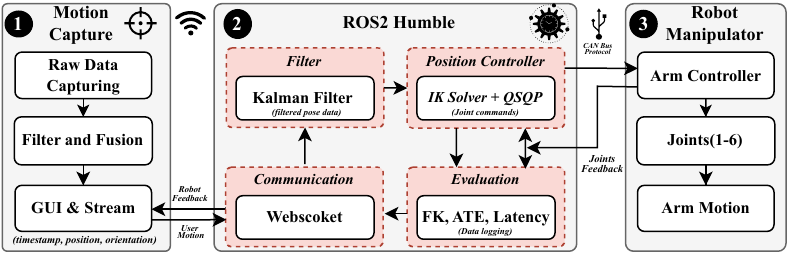}
    \caption{System Architecture. 
    \textnormal{The pipeline captures human intent via multimodal interfaces, processes it through a QP safety controller, and executes it on the Robotic manipulator, ensuring high-fidelity reproduction of user motion.}}
    \Description{A system architecture diagram illustrating a three-stage pipeline. The first stage is 'Capture,' showing inputs from ARPose and CV+IMU. The second stage is 'Control,' featuring a Newton-Raphson IK Solver and a QP Safety Filter. The final stage is 'Replay,' showing the command integration sent to the robotic manipulator.}
    \label{fig:sys_arch}
    \vspace{-0.2cm}
\end{figure*}

\section{System Design}
\label{sec:design}

To enable high-fidelity AR evaluation, we design \sysname as a transparent \emph{physical proxy} that captures natural human motions and accurately replicates them on a robotic manipulator. 
In this section, we discuss how we design the software framework of \sysname to achieve two conflicting goals: the \textit{freedom} of natural motion and the \textit{repeatability} of robotic execution. To demonstrate the versatility of our position-based control approach, we validated the system on both the OpenManipulator-X and the PiPER 6-DOF Arm; however, the comprehensive user study and dataset collection presented in this paper were conducted exclusively on the PiPER platform.

\subsection{Capturing Interfaces}
While recent works have explored AR interfaces for direct manipulation \cite{smith2025augmentedrealityinterfaceteleoperating} and generalized vision-based teleoperation \cite{qin2023anyteleop}, our platform focuses specifically on capturing high-frequency human motion ground truth for AR evaluation. Unlike AnyTeleop~\cite{qin2023anyteleop}, which prioritizes generalization across robot hands, \sysname requires a specialized, low-latency pipeline to capture high-velocity human dynamics (e.g., sudden tremors or rapid turns). 
Moreover, another design goal of \sysname is to allow researchers and practitioners to operate and collect additional AR user interaction data themselves in the field. As such, the human motion capturing methods should also be user-friendly and easy to use. 
We designed and implemented the following two methods. 

\paragraph{\textbf{ARPose (Mobile Interface).}} 
To provide the operator of \sysname, e.g., the AR user, a natural and intuitive method to interact with \sysname, we design an Android application that leverages ARCore's Visual-Inertial Odometry (VIO)~\cite{arcore} to transform a smartphone into a source of 6DOF position data. 
This allows \sysname to capture the AR user movement while the user directly interacts with a mobile handheld AR device. 
The 6DoF data, depicting the mobile device's poses, will also be used to control the robot's end-effector movement. 

\paragraph{\textbf{CV+IMU (Touchless Interface).}} To capture empty-handed interaction, we developed a hybrid tracking system combining a depth camera (Intel RealSense) with a wearable IMU. Unlike purely vision-based approaches that may lose tracking during fast motion, our sensor fusion algorithm uses the 200Hz IMU stream to maintain rotational fidelity, ensuring that even subtle hand micro-movements are preserved for the replay stage. Additionally, this interface supports an autopilot mode that drives the manipulator through standardized geometric trajectories (e.g., circles, squares) with sub-centimeter precision, enabling reproducible benchmarking without human input. This mode allows researchers to generate synthetic datasets of ideal motion, providing great baseline for evaluating tracking drift against known ground truth.

\subsection{Control Architecture}
\label{sec:control_arch}
To bridge the gap between noisy human input and precise robotic actuation, we developed a \textbf{position-centric proactively-safe architecture}.  While methods like RelaxedIK \cite{Rakita2018RelaxedIKRS} utilize non-linear optimization to synthesize feasible motion, we adopt shared autonomy principles similar to robotic nursing assistants \cite{boguslavskii2023}, our QP controller ensures that the robot faithfully executes the user's geometric intent while strictly adhering to hardware safety limits. This pipeline runs at 200Hz and provides both safety and smoothness guarantees. The control logic follows a three-stage process:

\paragraph{\textbf{Geometric Solver (Newton-Raphson)}} First, the system solves the Inverse Kinematics (IK) problem to find the ideal joint configuration $q_{\text{target}}$ that matches the user's hand pose. We use an iterative Newton-Raphson solver with Damped Least-Squares (DLS) to handle singularities and ensure numerical stability. Unlike analytical solvers that may oscillate near joint limits, the DLS method modifies the standard Jacobian pseudo-inverse update rule to penalize large joint velocities. The update step $\Delta q$ is computed as:
    \begin{equation}
        \Delta q = J^T (J J^T + \lambda^2 I)^{-1} \vec{e}
        \label{eq:dls}
    \end{equation}
    where $J$ is the geometric Jacobian, $\vec{e}$ is the task-space error vector between the current and desired end-effector pose, $I$ is the identity matrix, and $\lambda$ is a non-zero damping factor. This formulation ensures that the robot always finds a valid configuration close to the target without exhibiting erratic behavior when the Jacobian becomes ill-conditioned.
\begin{figure*}[t]
    \centering
    \begin{minipage}{0.45\textwidth}
        \centering
    \includegraphics[width=\linewidth]{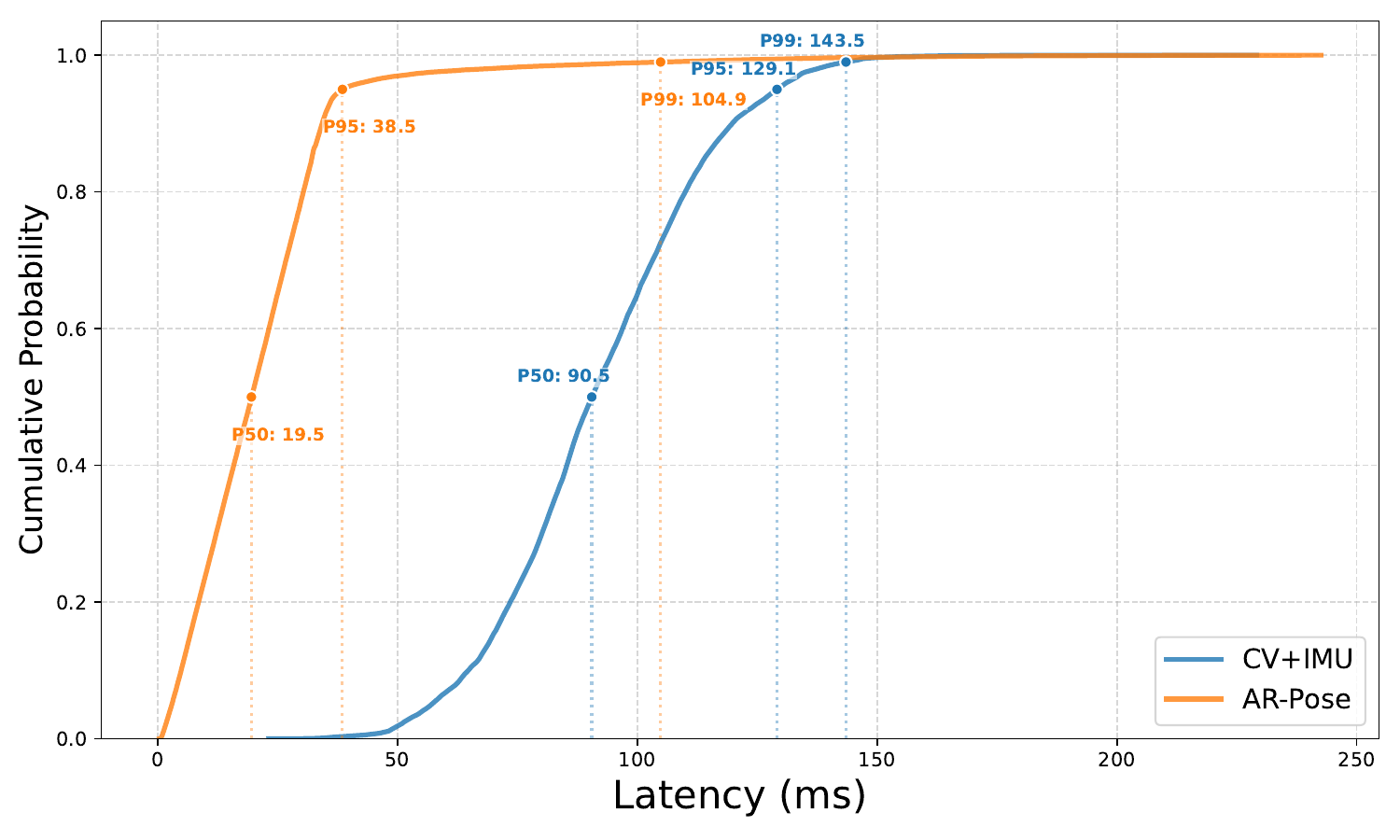}
     \caption{System Latency CDF. 
        \textnormal{ARPose interface shows consistently lower latency ($<40$ms), while CV+IMU exhibits a long tail due to processing overhead.}}
        \label{fig:cdf_latency}
        \Description{A Cumulative Distribution Function (CDF) line plot comparing system latency. The x-axis represents latency in milliseconds, and the y-axis represents probability. The ARPose curve rises sharply, indicating consistently low latency around 20ms. The CV+IMU curve has a shallower slope, indicating higher variable latency centered around 90ms.}
    \vspace{-0.3cm}
    \end{minipage}
    \hfill
    \begin{minipage}{0.45\textwidth}
        \centering
        \includegraphics[width=\linewidth]{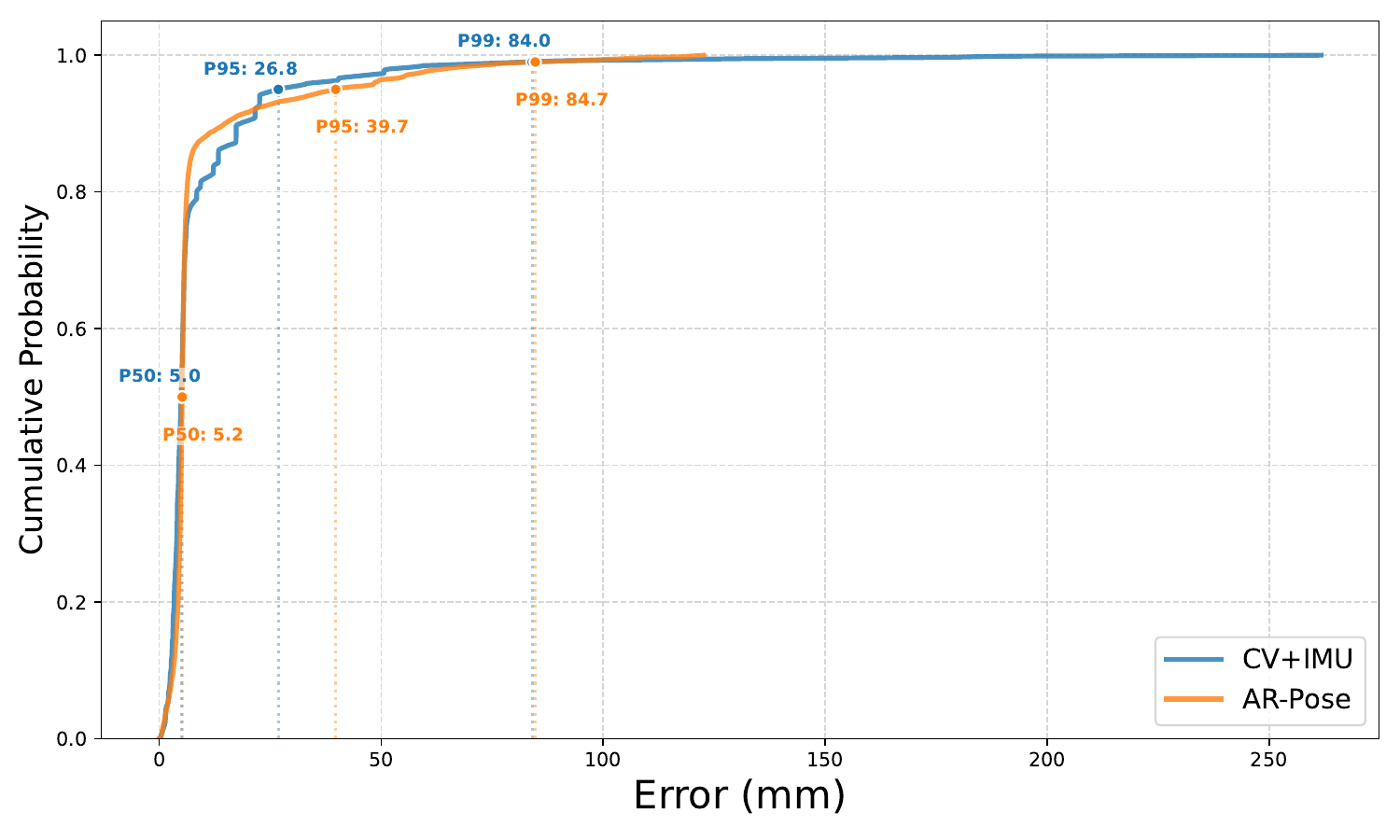}
        \caption{Tracking Error (ATE) CDF. 
        \textnormal{Both methods achieve ~5mm median accuracy, but CV+IMU shows better stability (P95) against tracking outliers.}}
        \Description{A Cumulative Distribution Function (CDF) plot showing Absolute Trajectory Error (ATE) in millimeters. Both ARPose and CV+IMU methods start near 0, with median values around 5mm. The CV+IMU curve reaches 1.0 (100 percent) faster than the ARPose curve, indicating fewer large error outliers for the CV+IMU method.}
        \label{fig:cdf_ate}
    \vspace{-0.3cm}
    \end{minipage}
\end{figure*}

\paragraph{\textbf{Safety Filter (Quadratic Programming).}} Once a target configuration is found, we do not command it directly. Instead, we formulate a Convex Optimization problem to compute the optimal velocity vector $\dot{q}^*$ that moves the robot toward $q_{target}$. We use a Quadratic Programming (QP) solver (OSQP) \cite{8516834} to minimize the tracking error subject to dynamic constraints:
    \begin{equation}
        \min_{\dot{q}} \quad || \dot{q} - \dot{q}_{needed} ||^2 \quad \text{s.t.} \quad \dot{q}_{min} \leq \dot{q} \leq \dot{q}_{max}
    \end{equation}
    where $\dot{q}_{needed}$ is the velocity required to reach the target in one time-step. This acts as a proactive safety filter, mathematically guaranteeing that the output command never violates velocity or acceleration limits, even if the user makes a sudden, erratic movement.
    
\paragraph{\textbf{Command Integration (Position Output).}} 
    Finally, we integrate the safe velocity $\dot{q}$ computed by the QP solver to generate the immediate next position command:
    \begin{equation}
        q_{\text{cmd}} = q_{\text{current}} + (\dot{q} \cdot \Delta t)
    \end{equation}
    This $q_{\text{cmd}}$ is sent to the robot's low-level position controller.

We chose this position-centric approach over standard velocity-based teleoperation to handle network instability. Our preliminary experiments revealed that velocity controllers suffer from Integration Drift during packet loss. If a stop command is lost in transmission, a velocity-controlled robot continues moving (Zero-Order Hold), potentially causing collisions. In contrast, our position-centric architecture creates a magnetic pull toward the last known valid pose. If the network drops, the robot simply stops at the target, inherently failing safe without requiring a watchdog timer.

\subsection{Implementation}
\label{sec:implementation}

The design described above is realized through a distributed software stack optimized for low-latency communication.

\paragraph{\textbf{Coordinate Homogenization}}
A critical engineering challenge was reconciling the coordinate systems of consumer AR devices with industrial robotics standards. ARCore utilizes a right-handed Y-up coordinate system (where $+Y$ opposes gravity), whereas ROS utilizes a Z-up convention. 
To prevent user disorientation, we implement a real-time homogenization layer on the client side. Before transmission, raw AR poses $P_{ar}$ are rotated into the robot's base frame using a fixed quaternion transformation $Q_{fix} = [0.707, 0, -0.707, 0]$. This ensures that forward motion of the phone maps correctly to forward motion of the robot arm, regardless of the user's initial orientation.

\paragraph{\textbf{Low-Latency Network Stack}}
To achieve our target end-to-end latency of $<95$ms, we implemented a dual-protocol network stack that minimizes serialization overhead.
For the high-frequency IMU data (200Hz), we use a custom binary protocol over serial. Packets are structured with a minimal 22-byte payload: a 1-byte header (\texttt{0xAA}), a 4-byte timestamp, and four 4-byte floats for the quaternion. This eliminates the parsing overhead of text-based formats like JSON.
For the ARPose application, we utilize a persistent WebSocket connection. While slightly heavier than UDP, WebSockets provide the reliability required for safety-critical commands. We optimize throughput by using a Drop-Oldest buffer policy on the receiver, ensuring the control loop always processes the freshest available packet rather than processing a queue of stale commands.

\paragraph{\textbf{Evaluation Package}}
To validate system performance rigorously, we developed a dedicated ROS2 package that runs in parallel with the control loop. The nodes subscribe to two synchronized data streams: the \textit{Target Pose} (user intent) and the \textit{Actual End-Effector Pose} (robot state). 
By utilizing message filters with approximate time synchronization, the ROS2 package pairs these streams to calculate the Tracking Error (RMSE) and End-to-End Latency in real-time. This instrumentation allows us to generate the precise numerical benchmarks presented in Section \ref{sec:evaluation}, confirming that the system maintains sub-centimeter accuracy.

\begin{figure*}[t]
    \centering
    \includegraphics[width=1.0\textwidth]{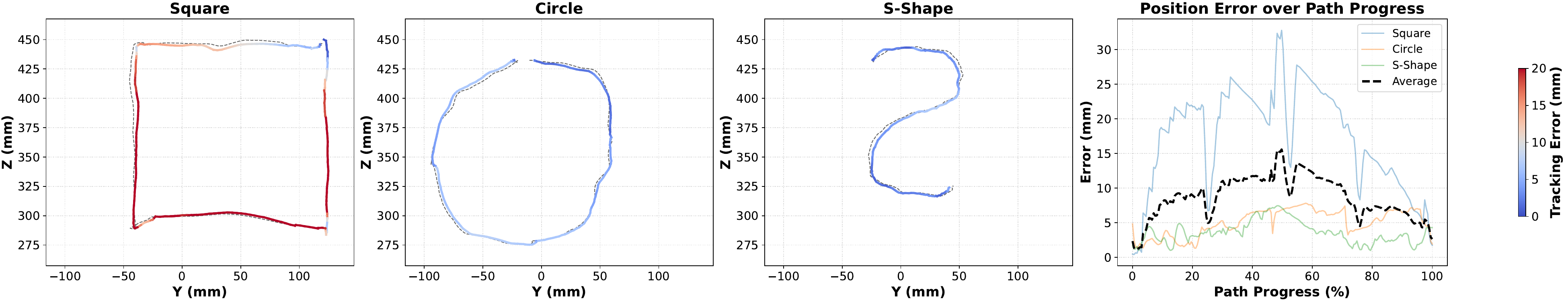}
    \caption{Spatial Dynamics Analysis. 
    \textnormal{Columns 1-3 display the error heatmaps for Square, Circle, and S-Shape trajectories. Blue indicates low error ($<7.5$mm) and Red indicates higher error ($<20$mm). Column 4 shows the temporal position error evolution.}}
    \Description{A set of spatial error heatmaps for three shapes: Square, Circle, and S-Shape. The trajectories are colored from blue (low error, less than 7.5mm) to red (high error, up to 20mm). The Square shape shows distinct red hotspots at the 90-degree corners, while the Circle and S-Shape are predominantly blue, indicating smooth tracking.}    
    \label{fig:spatial_dynamics}
    \vspace{-0.3cm}
\end{figure*}

\section{Evaluation}
\label{sec:evaluation}

We conducted a two-part evaluation of \sysname. First, an IRB-approved human-subject user study ($N=11$) was performed to assess the usability, workload, and precision of the two input modalities (CV+IMU vs. ARPose). Second, we leveraged the dataset collected during this study to analyze \sysname's tracking fidelity, latency, and repeatability.

\subsection{Study Protocol}
The study was conducted under an IRB-approved protocol. Participants were tasked with teleoperating the manipulator to trace specific geometric shapes while receiving real-time visual feedback of their trajectory relative to the target path on a computer monitor.

\subsubsection{Participants}
We recruited 11 participants (8 male, 3 female) from the university community, aged 21--29. The group consisted primarily of engineering graduate students with varying levels of familiarity with robotics. Participants reported their experience on a 5-point Likert scale: Gaming experience ($M=3.36, SD=1.63$), AR familiarity ($M=2.90, SD=1.38$), and Remote Control (RC) experience ($M=2.81, SD=1.17$). All subjects provided informed consent before participation.

\subsubsection{Apparatus}
The experimental setup included the \sysname Robotic manipulator and a host workstation (Legion 5 Pro, Intel Core i9-14900HX, RTX 4060 GPU, 16GB RAM) running the ROS2 control stack. The two input interfaces were:
(i) \textbf{CV+IMU Mode}: An Intel RealSense L515 LiDAR camera for depth sensing and an MPU-9250 IMU worn on the user's wrist.
(ii) \textbf{ARPose Mode:} A Google Pixel 6 Pro running the ARPose application.

\subsubsection{Experimental Tasks}
\label{sec:tasks}
Participants performed two distinct tasks designed to evaluate different aspects of the control system.

\paragraph{Task 1: Geometric Shape Tracing.}
Users were asked to trace three shapes a square, a circle, and an s-shape as accurately as possible using both interaction modes (counterbalanced to avoid order effects). 
These shapes were selected to test specific control capabilities:
\begin{itemize}
    \item \textbf{Square:} Evaluates the system's ability to handle sharp discontinuities (90-degree corners) and sudden velocity changes.
    \item \textbf{Circle:} Tests the maintenance of constant velocity and continuous curvature.
    \item \textbf{S-Shape:} Assesses the tracking of complex, free-form curves typical of natural human motion.
\end{itemize}

\paragraph{Task 2: Repeatability Test.}
To validate our hypothesis that robotic playback provides superior repeatability compared to human motion, we conducted a two-phase repeatability experiment. First, participants were asked to trace a rectangle shape three times in succession ($H_1, H_2, H_3$). Subsequently, the robot autonomously replayed their first attempt ($H_1$) five times ($R_1, \ldots, R_5$). We quantify repeatability using \textit{Inter-Trial Variability (ITV)} the mean standard deviation of position at corresponding time-aligned waypoints.

\subsubsection{Human Motion Dataset}
To support further research in AR interaction evaluation, we  logged the telemetry from all user trials. The resulting dataset consists of \textbf{132} total trajectories collected at 100Hz, organized into two subsets:
(i) \textbf{Subset A (Shape Tracing):} 66 trajectories from Task 1 (11 Users $\times$ 3 Shapes $\times$ 2 Modes). This subset covers distinct geometric primitives (Square, Circle, S-Shape) for accuracy analysis and forms the basis for the system performance analysis in \S\ref{sec:sys_perf}.
(ii) \textbf{Subset B (Repeatability):} 66 trajectories from Task 2 (11 Users $\times$ 3 Rectangle Trials $\times$ 2 Modes). This subset captures the natural variability of human motion over repeated attempts at the same task.

Each log file (CSV) contains time-synchronized streams of target human pose captured via CV+IMU or ARPose, and the actual kinematic pose executed by the manipulator. Beyond raw trajectory data ($p_x, p_y, p_z, q_x, q_y, q_z, q_w$), the dataset includes critical system performance metrics for every frame, including Inverse Kinematics (IK) solver delay, end-to-end latency, and synchronization deltas. This comprehensive logging enables researchers to analyze not just the spatial accuracy of the motion, but also the temporal characteristics of the teleoperation pipeline.

\subsection{User Analysis}
\label{sec:user_analysis}

\subsubsection{Workload (NASA-TLX)}
After interacting with each capture mode, participants completed a NASA-TLX survey to rate their perceived workload. Interestingly, the reported workload was nearly identical between conditions, with the ARPose interface ($M=28.4, SD=16.5$) scoring similarly to the CV+IMU interface ($M=28.6, SD=11.6$). This suggests that despite the physical difference of holding a device versus free-hand motion, the cognitive load of the teleoperation task remained consistent, validating the intuitive nature of both mapping algorithms.

\subsubsection{Usability (SUS)}
The System Usability Scale (SUS) scores reflect a high degree of acceptance for both systems. The ARPose interface achieved a mean SUS score of \textbf{81.1} ($SD=13.7$), corresponding to an A grade. The CV+IMU interface followed closely with a mean score of \textbf{79.5} ($SD=14.9$), a B+ grade. These results indicate that the touchless CV interface, often considered harder to control, was implemented with sufficient robustness to rival the familiarity of a touchscreen interface.

\subsubsection{User Preference}
In post-study interviews, 7 out of 11 participants preferred the ARPose interface, citing the tactile feedback of the phone and lower perceived latency. The remaining 4 participants preferred the CV+IMU mode, highlighting the benefit of hands-free operation and the natural 1:1 wrist rotation mapping.

\subsection{System Performance Analysis}
\label{sec:sys_perf}
We evaluated the system's technical performance using the \textbf{66-trajectory dataset} collected in Task 1. The analysis focuses on two primary metrics:
(i) \textbf{System Latency:} The end-to-end time elapsed from sensor data capture to robot actuation command.
(ii) \textbf{Tracking Error (ATE):} The Root Mean Square Error (RMSE) between the commanded trajectory and the robot's actual executed path.

\begin{figure*}[t]
    \centering
    \includegraphics[width=\linewidth]{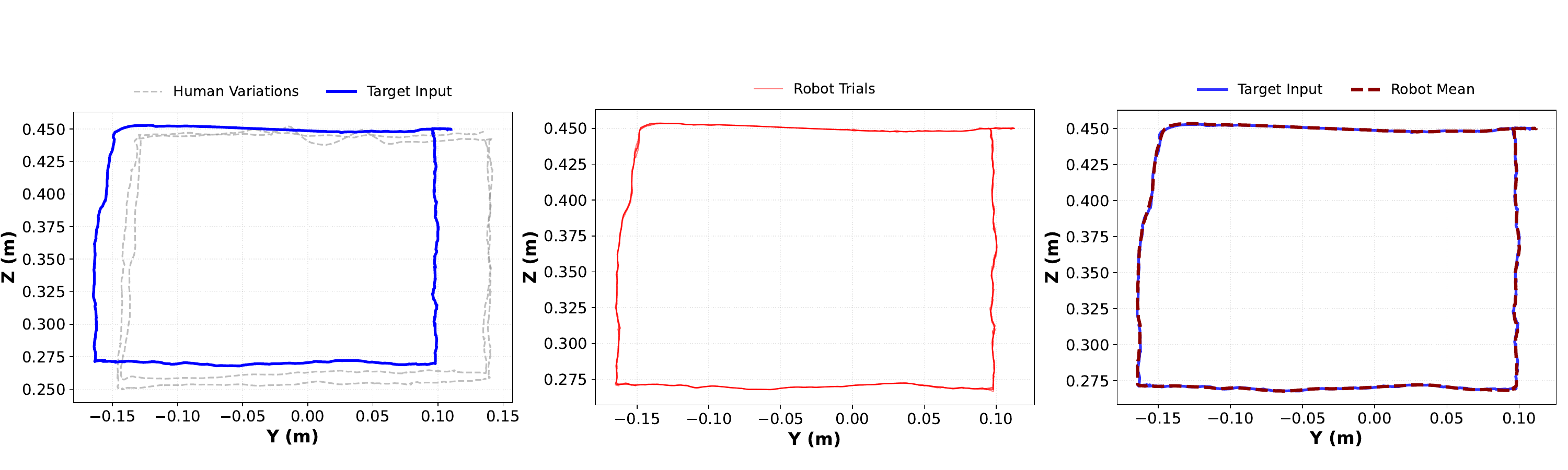}
    \caption{Repeatability Comparison. 
    \textnormal{Left: Three human trials showing significant natural variation (ITV $\approx$ 27.69mm). Center: Five robot replays mimicking one human trajectory. Right: Five robot replay trials of the recorded motion, showing near-perfect overlap (ITV $<$ 3.91mm).}}
    \label{fig:repeatability}
\Description{A side-by-side comparison of 3D trajectory paths. The left plot shows three human trials of a rectangular shape, exhibiting visible spatial variation and jitter (labeled Human ITV 27.69mm). The right plot shows five robotic replay trials of the same shape, where the lines overlap nearly perfectly with minimal variation (labeled Robot ITV 3.91mm).}
\end{figure*}

\subsubsection{Latency Analysis}
Figure~\ref{fig:cdf_latency} shows the CDF of system latency for each of the two modalities. The \textbf{ARPose mode} demonstrates superior responsiveness, with a median latency (P50) of \textbf{19.5 ms}. This sharp rise in the CDF implies that the ARPose pipeline which relies on lightweight UDP/WebSocket transmission of pre-computed trajectory imposes minimal computational overhead.
In contrast, the \textbf{CV+IMU mode} shows a broader distribution with a median latency of \textbf{90.5 ms}. This observation suggests that the heavy lifting of depth map processing and skeletal inference on the host PC is the primary bottleneck. While 90 ms is perceptible, the smooth slope of the CDF indicates that this latency is consistent, which is critical for predictability in teleoperation.

\subsubsection{Tracking Fidelity (ATE)}
Figure~\ref{fig:cdf_ate} illustrates the distribution of Absolute Trajectory Error (ATE). Both systems achieve a comparable median tracking error of approximately \textbf{5.0 mm}. This sub-centimeter accuracy validates the efficacy of the underlying QP controller in faithfully executing commands from either source.
However, the tail of the distribution reveals a key insight: the \textbf{CV+IMU mode} outperforms ARPose in worst-case stability (P95 ATE of 26.8 mm vs. 39.7 mm). The stepped behavior in the ARPose curve's tail suggests occasional drift events, likely due to VIO losing tracking features during rapid device motion. Conversely, the CV+IMU system, which fuses absolute depth readings with 200Hz IMU data, is more robust to these transient tracking failures.

\subsection{Trajectory Analysis}
\label{sec:traj_analysis}
To further understand how the system handles different motion primitives, we analyzed the spatial error distribution across the three shapes (Square, Circle, S-Shape) using the heatmaps shown in Figure~\ref{fig:spatial_dynamics}.
The \textbf{Square} trajectory (Column 1) reveals error hotspots (red) concentrated strictly at the 90-degree corners. This confirms that the QP safety filter is actively intervening to limit acceleration; as the user abruptly changes direction, the robot smooths the corner to prevent jerk. The \textbf{Circle} and \textbf{S-Shape} trajectories (Columns 2 and 3) exhibit a remarkably uniform blue error distribution, implying that for continuous, fluid motions, the controller introduces negligible distortion.

\subsection{Repeatability Analysis}
\label{sec:repeatability}
A core feature of \sysname is its ability to serve as a high-fidelity record and replay proxy. We evaluate \sysname's repeatability by comparing the Inter-Trial Variability (ITV) of human motion to the robot's replay of that same motion.
Figure~\ref{fig:repeatability} illustrates this comparison using a representative trial from User 4 with ARPose. We observe that natural human motion varies significantly between attempts, exhibiting a spatial spread (Human ITV) of 27.69 mm. However, when the robot replays one of these recorded trajectories (right), the variability drops to just 3.91 mm, effectively filtering out the stochastic nature of human motor control while preserving the original trajectory's intent.
This trend holds true across the entire dataset ($N=11$). For the ARPose mode, the average human ITV was 75.59 mm, whereas the robot ITV was 7.40 mm, a 10.2$\times$ improvement in consistency. Similarly, for the CV+IMU mode, the average human ITV was 46.48 mm compared to that of \sysname at 13.73 mm. These results show that \sysname can accurately replay user motion, enabling researchers to run repeatable benchmarks on AR tracking algorithms.

\section{Conclusion}

Current AR evaluation methodologies suffer from a critical lack of physical ground truth, relying often on software simulations or inconsistent human testing. To address this, we presented \sysname, a robotic platform designed to serve as a precise, programmable proxy for human motion. Our evaluation shows that \sysname can achieve sub-centimeter tracking accuracy and consistent system latency.
By open sourcing \sysname, we hope to provide the community with a robust hardware-in-the-loop testbed that facilitates human-centered AR evaluations.

\section{ACKNOWLEDGMENTS}
This work was supported in part by NSF Grants \#2236987, \#2346133, and \#2402383. We thank Harsh Shah for assistance with final code cleanup and preparing the public GitHub release.
\bibliographystyle{ACM-Reference-Format}


\end{document}